\def\year{2020}\relax
\documentclass[letterpaper]{article} 
\usepackage{aaai20}  
\usepackage{times}  
\usepackage{helvet} 
\usepackage{courier}  
\usepackage[hyphens]{url}  
\usepackage{graphicx} 
\usepackage{graphicx}  
\nocopyright

\usepackage{booktabs}
\usepackage{amsfonts} 
\usepackage{multirow,array}

\title{Brain-mediated Transfer Learning of Convolutional Neural Networks}
\author{Satoshi Nishida,\textsuperscript{\rm 1} Yusuke Nakano,\textsuperscript{\rm 1} Antoine Blanc,\textsuperscript{\rm 1} Naoya Maeda,\textsuperscript{\rm 2} \\ \Large \textbf{Masataka Kado,}\textsuperscript{\rm 2} \textbf{Shinji Nishimoto}\textsuperscript{\rm 1} \\ 
\textsuperscript{\rm 1}Center for Information and Neural Networks (CiNet), \\ National Institute of Information and Communications Technology (NICT), Osaka, Japan \\ 
\textsuperscript{\rm 2}NTT DATA Corporation, Tokyo, Japan \\
\{s-nishida, nakano, blanc\}@nict.go.jp, \{naoya.maeda, masataka.kado\}@nttdata.com, nishimoto@nict.go.jp 
}
 
\begin{document}

\maketitle

\begin{abstract}
The human brain can effectively learn a new task from a small number of samples, which indicate that the brain can transfer its prior knowledge to solve tasks in different domains. This function is analogous to transfer learning (TL) in the field of machine learning. TL uses a well-trained feature space in a specific task domain to improve performance in new tasks with insufficient training data. TL with rich feature representations, such as features of convolutional neural networks (CNNs), shows high generalization ability across different task domains. However, such TL is still insufficient in making machine learning attain generalization ability comparable to that of the human brain. To examine if the internal representation of the brain could be used to achieve more efficient TL, we introduce a method for TL mediated by human brains. Our method transforms feature representations of audiovisual inputs in CNNs into those in activation patterns of individual brains via their association learned ahead using measured brain responses. Then, to estimate labels reflecting human cognition and behavior induced by the audiovisual inputs, the transformed representations are used for TL. We demonstrate that our brain-mediated TL (BTL) shows higher performance in the label estimation than the standard TL. In addition, we illustrate that the estimations mediated by different brains vary from brain to brain, and the variability reflects the individual variability in perception. Thus, our BTL provides a framework to improve the generalization ability of machine-learning feature representations and enable machine learning to estimate human-like cognition and behavior, including individual variability.
\end{abstract}

\section{Introduction}

\noindent Transfer learning (TL) is an effective framework that improves the generalization ability of machine learning in pattern recognition \cite{pan2009}. In TL, feature representations trained with large-scale, high-quality data in a specific task are transferred to other tasks in different domains with insufficient training data. Recently, TL with convolutional neural networks (CNNs) has received increasing attention owing to its high generalization ability across various task domains \cite{tan2018}. Despite the fact that the training of CNNs requires massive amounts of labeled datasets, CNNs acquire generic feature representations available for various types of pattern recognition through TL \cite{cimpoi2016,donahue2014,girshick2014,oquab2014,razavian2014,toshev2014,xiao2014}. For example, Donahue et al. (2014) used a CNN feature space, pre-learned with an object recognition dataset, for tasks different from the original challenge, such as scene classification. Using higher-layer features of the pre-trained CNN, they trained classifiers for those new tasks and achieved state-of-the-art performance. In this way, TL improves the generalization ability of machine learning more effectively when using more generic feature representations. However, the generalization ability of machine learning is still much lower than that of the human brain.

The human brain has a sophisticated ability to generalize the knowledge acquired through limited experiences to cognition and behavior under novel situations. This indicates that the human brain can effectively use its own internal representations of cognitive information across different task domains. Furthermore, such representations can be used as a medium of brain decoding that infers various types of cognition from brain activity, such as semantic cognition \cite{mitchell2008,nishida2018}, affections \cite{kim2015,peelen2010}, economic decision \cite{hampton2007,knutson2007}, human mass behavior \cite{dmochowski2014,falk2012}. Therefore, machine-learning feature representations may become more generic by incorporating the brain representations into them, leading to further improvements in the generalization ability of machine learning through TL.

Recently, several studies have proposed techniques to guide pattern recognition with CNNs by combining CNN features with human brain activity \cite{fong2018,spampinato2017}. For example, Fong et al. (2018) trained a classifier of images using higher-layer features of a CNN while guiding the training by voxel responses to the images measured using functional magnetic resonance imaging (fMRI). To make the decision surface of the classifier more consistent with brain representations, the training was weighted by voxel responses. They demonstrated higher performance in image recognition for the classifier with voxel response weighting than for the classifier without it. These studies suggest that CNN feature representations can improve by combining them with human brain activity. However, the recognition task they targeted was image recognition, in which CNNs had already shown splendid performance by themselves. In addition, when their techniques are applied to other types of recognition tasks, they require brain data dedicated to each of the tasks. Therefore, whenever their techniques are applied to a novel task, new measurements of brain data specified to that task are required. These limitations prevent their techniques from improving the generalization ability of machine learning.

To address these issues, we propose, in this paper, brain-mediated TL (BTL). In the BTL, the voxel response data measured from the human brain in an fMRI experiment are combined with CNN features to be used for various types of recognition tasks. First, CNN features for audiovisual inputs are transformed into features of the individual brains using a pre-learned linear mapping between them. Then, the transformed features are used to train learners that estimate arbitrary pairs of audiovisual inputs and corresponding labels. We apply this BTL to the estimation of labels to which human subjective cognition is strongly related. Our results demonstrate that the BTL shows higher estimation performance than regular TL and the brain-by-brain variation of estimation by the BTL reflects individual variability in subjective cognition.

\section{Model}

\subsection{Procedure}

\begin{figure*}
\centering
\includegraphics[width=0.95\textwidth]{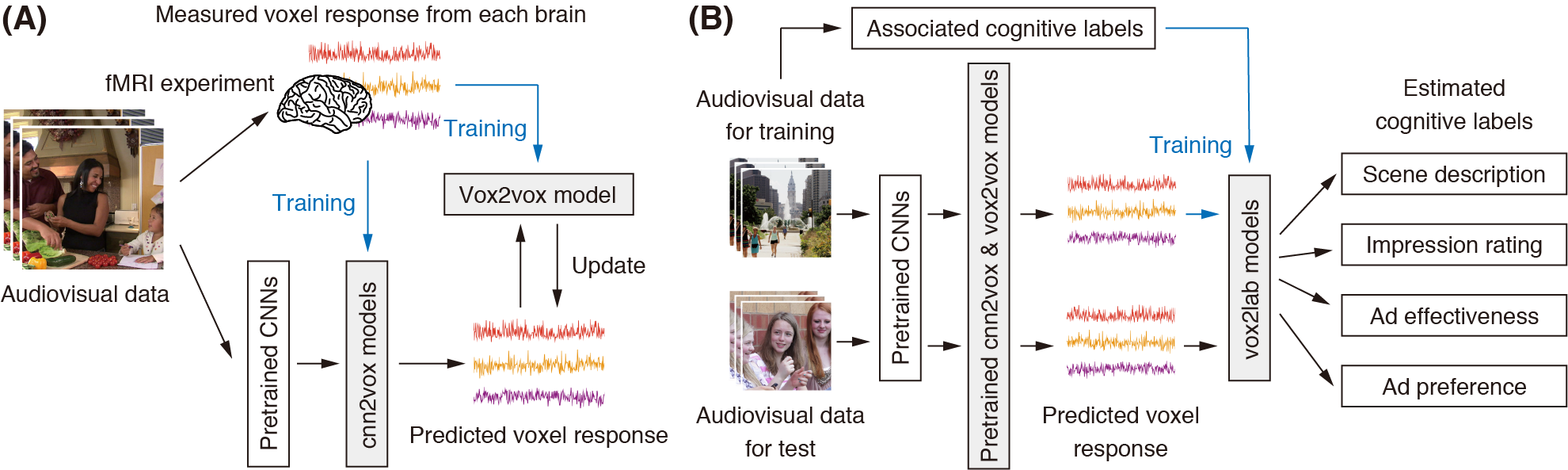} 
\caption{A schematic of the proposed brain-mediated transfer learning (BTL) method. (A) Voxel-response prediction from audiovisual inputs. (B) Label estimation using predicted response.}
\label{fig:schematic}
\end{figure*}

Our modeling procedure for BTL is divided into the four steps (Figure \ref{fig:schematic}): (1) the feature extraction from the inputs of movies and/or sounds via CNNs, (2) the prediction of voxel responses to the inputs using models that linearly transform the extracted features to the voxel responses (CNN-to-voxel [cnn2vox] models), (3) the modification of the predicted responses using models that predict voxel response from the history of preceding voxel responses (voxel-to-voxel [vox2vox] models; Figure \ref{fig:schematic}A), and (4) the estimation of cognitive labels from the predicted responses using models that have linear association between them (voxel-to-label [vox2lab] models; Figure \ref{fig:schematic}B).

The cnn2vox and vox2vox models are pretrained using small datasets of movie-evoked voxel responses collected from individual brains with fMRI. Once the training is done, these models predict voxel responses to arbitrary audiovisual inputs by transforming the CNN features to the response of individual brains with no additional brain measurement. Supposing that voxel responses convey brain feature representations of the inputs, this process corresponds to the transformation from CNN features to brain feature representations. Then, the vox2lab model is trained using paired datasets of predicted voxel response and cognitive labels linked with audiovisual inputs. Since this training does not require datasets used in fMRI experiments, our method allows acquiring the association between arbitrary pairs of audiovisual inputs and cognitive labels through the feature transformation.

\subsection{CNN Feature Extraction}

In this study, VGG-16 \cite{simonyan2014} and SoundNet \cite{aytar2016}, which are pretrained and released on the web, are used to extract visual and acoustic features, respectively, from movies. To extract visual features from movies via VGG-16, which was originally applied to static images with a fixed size of 224 $\times$ 224 pixels, the movies are decomposed into frames and resized to the same size. Then, unit activations of intermediate layers when inputting the movie frames are calculated and pooled for each second. Finally, the maximum activation value of each unit for each second is used as the visual features of the movies. This study uses eight layers of pool1--5 and fc6--8 and obtains the visual features for each layer. To extract acoustic features from the same movies, the sound waves of the inputs are resampled with the fixed frequency of 44100 Hz and decomposed into each second. Then, unit activations of intermediate layers in SoundNet when inputting the sound waves are calculated as the acoustic features of the movies. This study uses seven layers of conv1--5 and fc 6--7 and obtains the acoustic features for each layer. Finally, these processes produce eight series of visual features and seven series of acoustic features from time series of movies.

\subsection{Cnn2vox Model}

The construction of the cnn2vox, vox2vox, and vox2lab models is based on the voxelwise modeling technique \cite{naselaris2011}. Using a time series of features and voxel responses, the cnn2vox model acquires the linear mapping from a CNN feature space to a response space of each voxel through statistical learning. The learning objective is to estimate weights of $N$ voxels, denoted by ${\bf W}_{\rm cv} = \{{\bf w}_{\rm cv(1)}, \cdots, {\bf w}_{{\rm cv}(N)}\}$, of the linear model: ${\bf R}=f({\bf X}) {\bf W}_{\rm cv}+\epsilon$, where ${\bf R}=\{{\bf r}_1, \cdots, {\bf r}_N\}$ is a series of responses in each of $N$ voxels, $\bf X$ is a series of audiovisual inputs, $f({\bf X})$ is its feature representation with the dimensionality of $D$, and $\bf \epsilon$ is isotropic Gaussian noise. A set of linear temporal filters is used to capture the hemodynamic delay in the response \cite{nishimoto2011}. The matrix of $f({\bf X})$ is constructed by concatenating four sets of $D$-dimensional feature vectors with temporal shifts of 3, 4, 5, and 6 s. This means that voxel response at a time point $t$, denoted by ${\bf R}_{(t)}~(t=1,\cdots,T)$, is modeled by a weighted linear combination of the preceded series of features:
\[ {\bf R}_{(t)}=\sum_{(k=3,4,5,6)}f({\bf X}_{(t-k)}) {\bf W}_{{\rm cv},k}+{\bf \epsilon}, \]
where ${\bf W}_{{\rm cv},k}$ denotes the weights corresponding to the delay $k$. The weight estimation is performed using L2-regularized linear least-squares regression. The optimal regularization parameter for each model is determined by 10-fold cross validation of training data and shared across all voxels. In this study, $f({\bf X})$ represents a series of unit activations induced by movies for each layer of VGG-16 or SoundNet. Since the very large number of units in lower layers of the CNNs took too much computational cost for the regression process, the dimensionality of unit-activation features for each layer is reduced in advance by principal component analysis (PCA) on training datasets. This study reduces the dimensionality, $D$, to 1000. Finally, eight models for eight VGG-16 layers (pool1--5 and fc6--8) and seven models for seven SoundNet layers (conv1--7) are constructed for each brain.

Each of the estimated linear models predicts voxel responses to new movie inputs. Then, the predicted voxel responses from individual models are integrated using linearly weighted average for each voxel. The weight for a given voxel is calculated based on the prediction accuracy (Pearson's correlation coefficient between measured and predicted voxel responses) for that voxel calculated during the cross-validation in model training. In particular, the weight for the $i$-th model, denoted by $w_i$, is determined by $w_i=a_i/\sum_j^{15}a_j$ , where $a_i$ is the prediction accuracy of the model. This integration process produces a single series of predicted voxel responses to the new movie inputs.

\subsection{Vox2vox Model}

The vox2vox model predicts response in one voxel at a given time point from responses in a group of voxels at the preceding time points. Hence, this model captures endogenous properties of voxel responses, such as intrinsic connectivity between different brain regions \cite{fox2007}, whereas the cnn2vox model captures exogenous properties of voxel responses, such as stimulus selectivity. Although the vox2vox model is not indispensable for BTL, modifying response prediction by the vox2vox model improves estimation performance in some cases.

In the vox2vox model, a response in each of $N$ voxels at a time point $t$, denoted by ${\bf R}_{(t)}$, is modeled by a weighted linear combination of responses in the set of $M$ voxels preceded by 1, 2, and 3 s:
\[ {\bf R}_{(t)}=\sum_{k=1,2,3}{\bf R}'_{(t-k)}  {\bf W}_{{\rm vv},k} + {\bf \epsilon}. \]
The $M$ voxels are selected on the basis of the cnn2vox model prediction accuracy on model training data. In this study, the top 2000 voxels with the highest prediction accuracy after the weighted average of all the models are used as the $M$ voxels. The regression procedure is the same as the one used in the cnn2vox model. Response predictions from the cnn2vox and vox2vox models are combined by weighted sum. The weight is determined by the relative accuracy of each prediction for each voxel.

\subsection{Vox2lab Model}

The vox2lab model estimates cognitive labels associated with audiovisual inputs from predicted voxel responses. In this model, a series of z-scored cognitive labels at a time point $t$, denoted by ${\bf L}_{(t)}~(t=1, \cdots, T')$, is regressed by a series of predicted responses to the inputs in the set of $N$ voxels with the hemodynamic delay, $k$, of 3, 4, and 5s:
\[ {\bf L}_{(t)}=\sum_{k=3,4,5} {\bf \hat{R}}_{(t+k)}  {\bf W}_{{\rm vl},k} + {\bf \epsilon}. \]
The regression procedure is the same as the one used in the other models, except that the regularization parameters are determined separately for each dimension of label vectors. This model learns the association between predicted voxel responses (but not measured voxel responses) and cognitive labels. This allows us to associate arbitrary pairs of audiovisual data and corresponding labels, even when they are not used for training either the cnn2vox or vox2vox models.

However, the regression occasionally requires too much computational cost when the paired data have a large number of samples. In such cases, the number of voxels used for the modeling is reduced in advance by PCA on the training datasets. Then, the top PCs are used for the training of the three models. In this study, the top 300 PCs are used as voxels to predict (i.e., $N = 300$) for only one estimation task (see below). In addition, the top 10 PCs with the highest prediction accuracy are used for the regressors of the vox2vox model (i.e., $M = 10$). Even in this case, the other parameters and the regression procedure remain the same.

\section{Data}

\subsection{Movie}

Two sets of movies were provided by NTT DATA Corp. (Tokyo, Japan). One includes 368 Japanese ad movies broadcasted on the web between 2015 and 2018 (web ad movies). The movies were divided into 7200 s and 1200 s to collect voxel responses for the training (training dataset) and test (test dataset), respectively, of all the three models. The other includes 2452 Japanese ad movies broadcasted on TV between 2011 and 2017 (TV ad movies). Out of them, only 420 movies were used to collect voxel responses for the training of the cnn2vox and vox2vox models. The remaining 2032 movies were not used for fMRI experiments but for the training of the vox2lab model and the test of its estimation performance. The movies are all unique, include a wide variety of product categories (Supplementary Table 1), and have the same resolution (1280 $\times$ 720 pixels) and frame rate (30 Hz). The length of the movies is typically either 15 or 30 s. They are also accompanied by PCM sounds with the sampling rate of 44100 Hz and are normalized so that they have the same RMS level.

\subsection{fMRI Data}

fMRI responses to the movies were collected from Japanese participants using a 3T MRI scanner. Forty and twenty-eight participants were assigned to fMRI experiments with the web ad movies and those with the TV ad movies, respectively. The experimental protocol is approved by the ethics and safety committees of NICT. For the modeling in each participant, the fMRI data were preprocessed and all voxels within the whole cortex were extracted. For more details, see Supplementary Methods.

\subsection{Cognitive Labels}

The four types of cognitive labels associated with the movie datasets are used for testing the performance of BTL and comparing it with the performance of other methods: (1) scene descriptions, (2) impression ratings, (3) ad effectiveness indices, and (4) ad preference votes. All the label sets reflect rich perceptual information and/or complex behavioral outcomes related to human subjective cognition. The label sets (1)--(3) are linked to the web ad movies, whereas the label set (4) is linked to the TV ad movies. The label sets (1), (3), and (4) were provided from NTT DATA Corp., whereas the label set (2) was collected on psychological experiments conducted in our lab. In the following, the details of each cognitive label set are described (for more details, see Supplementary Methods).

Scene description data were collected from human annotators as manual descriptions given for every 1-s scene of the web ad movies. They were instructed to describe each scene using more than 50 Japanese characters. The descriptions contain a variety of expressions reflecting not only their objective perceptions but also their subjective perceptions (e.g., impression, feeling, association with ideas). To evaluate the scene descriptions quantitatively, the descriptions were transformed into vectors of word2vec \cite{mikolov2013}. Individual words in each description were projected into the pretrained word2vec vector space. Then, the word vectors obtained from all descriptions within each scene were averaged. This procedure yielded one 100-dimensional vector for each 1-s scene.

Impression rating data were collected from manual ratings on 30 different impression items conducted by human annotators. The ratings were given for every 2-s scene of the web ad movies. While the annotators sequentially watched 2-s separate clips of the movies, they evaluated each item on a scale of 0 to 4.The mean impression ratings in every 2-s scene were obtained by averaging multiple ratings and then oversampled to obtain time series of rating labels in every 1-s scene.

Ad effectiveness index data for the web ad movies were obtained as two types of mass behavior indices collected on the web. One index is click rate, that is, the fraction of viewers who clicked the frame of a movie and jumped to a linked web page. The other index is view completion rate, that is, the fraction of viewers who continued to watch an ad movie until the end without choosing a skip option. Although a single value of each index was assigned to each ad, time series of indices in every 1-s scene were obtained by filling all scenes in an ad with an identical index value assigned to the ad.

Ad preference vote data for the TV ad movies were collected for commercial investigation using questionnaires to large-scale testers. Each tester was asked to freely recall a small number of her/his favorite TV ads from among the ads recently broadcasted. The total number of recalls of an ad was regarded as its preference value. Although one preference value was assigned to each ad, time series of the value in every 1-s scene were obtained by filling all scenes in an ad with an identical value assigned to the ad. Since the preference data are distributed in a similar form of a gamma distribution, the logarithm of the data was taken. Only for this dataset, due to the large number of samples for training the vox2label model in the dataset (2032 movies, 45375 s), the voxel dimensionality is reduced to 300 in advance by PCA for computational efficiency of the modeling (see also BTL).

\section{Reference Methods}

To compare the performance of label estimation with BTL, two other methods are implemented.

\paragraph{Regular TL.}

TL estimates cognitive labels without the transformation from CNN features to voxel responses. Although the CNN feature extraction using VGG-16 and SoundNet is the same as the one used in BTL, CNN features are directly used to regress cognitive labels. In particular, a label at a time point $t$, denoted by ${\bf L}_{(t)}$, is modeled by a weighted linear combination of CNN features of the input at the same time point. For the neutral comparison with BTL, the reduction of CNN feature dimension to 1000 and the regression is performed in the same manner as used in BTL.

In contrast to this model, the BTL models contain linear temporal filters that capture the hemodynamic response delay. This may enable only BTL to estimate labels from a series of features at multiple time points. If so, the comparison between BTL and TL becomes unfair. To avoid this, another TL model is also intorduced so that a label at a given time point is estimated by a combination of CNN features at $-1$, 0, and 1 s from that time point: ${\bf L}_{(t)}=\sum_{k=-1,0,1} f({\bf X}_{(t-k)}) {\bf W}_{{\rm TL},k} + {\bf \epsilon}$.

These TL models are constructed using each of the 15 layers in VGG-16 and SoundNet. The label estimation from each layer model for new movie inputs is integrated using linearly weighted average for each dimension of label vectors. The averaging weight of $i$-th model, denoted by $w_i$, is determined according to label estimation accuracy of the model during the cross-validation in the model training, denoted by $a_i$. In particular, the averaging weight is represented as $w_i=a_i/\sum_j^{15} a_j$. This integration process produces a single series of label estimation to the new movie inputs.

\paragraph{Brain Decoding (BD).}

BD is an effective method that estimates human perception and cognition induced by complex audiovisual inputs \cite{nishida2018}. In the BD of this study, cognitive labels are directly estimated by analyzing measured voxel responses to movies. The model form is the same as in the vox2lab model of BTL except that the measured voxel responses are used instead of the predicted responses. The regression procedure is the same as the one used in BTL.

\section{Results}

\subsection{Performance in Voxel-response Prediction}

To confirm that our models appropriately extract audiovisual features of individual brains, we first examined the performance of cnn2vox and vox2vox models in terms of the prediction of voxel responses to web ad movies (Table \ref{tab:predacc}). The prediction accuracy was evaluated by Pearson's correlation between the predicted and measured voxel responses in the test fMRI dataset, which have not been used for model training. Each accuracy was averaged over the whole cortex and across all participants.

\begin{table}
  \caption{Accuracy of voxel-response prediction}
  \label{tab:predacc}
  \centering
  \begin{tabular}{cccc}
    \toprule
    \multicolumn{2}{c}{Cnn2vox} \\
    \cmidrule(r){1-2}
    VGG-16 & SoundNet & Vox2vox & Prediction accuracy \\
    \midrule
    \checkmark &  &  & 0.096 \\
    \checkmark &  & \checkmark & 0.098 \\
     & \checkmark &  & 0.054 \\
     & \checkmark & \checkmark & 0.055 \\
    \checkmark & \checkmark &  & 0.106 \\
    \checkmark & \checkmark & \checkmark & 0.108 \\
    \bottomrule
  \end{tabular}
\end{table}

The accuracy is higher when the predictions from both VGG-16 and SoundNet cnn2vox models were integrated than when only either of them was used. This is due to the fact that the VGG-16 and SoundNet features distinctively model audiovisual responses of voxels in different cortical regions. Indeed, the VGG-16 and SoundNet models accurately predict voxel responses in visual and auditory areas, respectively (Supplementary Figure 1). This is consistent with the previous findings that CNNs trained by datasets of specific modality predict voxel responses in brain regions involved in the processing of that modality \cite{guclu2017,guclu2015,kell2018}. Thus, our cnn2vox models well capture different modality of feature representations in the cortex. In addition, combining the prediction from the vox2vox models with the prediction from the cnn2vox models slightly improves the accuracy (Table \ref{tab:predacc}), indicating that endogenous properties of voxel response represented by the vox2vox models augment the model ability to describe the brain feature representation by adding them to exogenous properties of voxel response.

\subsection{Label Estimation Tasks}

\paragraph{Task1: Estimation of Semantic Perception.}

The first task is the estimation of semantic perception induced by each movie scene using the scene description data. All possible combinations of the cnn2vox and vox2vox models were calculated for BT, whereas for TL, all possible combinations of VGG-16 and SoundNet were calculated. Furthermore, for a fair comparison with BTL, the TL with the estimation from multiple time points was also computed (see Reference Methods). The performance of BTL and BD was evaluated using the mean estimation derived from the estimation of each participant model. Out of all the methods, the performance of BTL with the VGG-16 cnn2vox model is highest (Table \ref{tab:estpfm}). The best performance in BTL is significantly higher than the best performance in TL and the performance in BD (bootstrapping test, $p < 0.0001$).

\begin{table*}
  \caption{Estimation performance in all the tasks}
  \label{tab:estpfm}
  \centering
  \begin{tabular}{lcccccccc}
    \toprule
     & \multicolumn{2}{c}{CNN} & & \multicolumn{5}{c}{Estimation performance} \\
    \cmidrule(r){2-3} \cmidrule(r){5-9}
     & VGG-16 & SoundNet & Vox2vox & Task1 & Task2 & Task3-1 & Task3-2 & Task4 \\
    \midrule
    \multirow{6}*{BTL} & \checkmark & & & \textbf{0.546} & 0.498 & 0.505 & 0.440 & 0.354 \\
     & \checkmark & & \checkmark & \textbf{0.546} & 0.500 & 0.502 & 0.444 & 0.360 \\
     & & \checkmark & & 0.146 & 0.234 & 0.094 & 0.223 & 0.244 \\
     & & \checkmark & \checkmark & 0.146 & 0.235 & 0.093 & 0.227 & 0.252 \\
     & \checkmark & \checkmark & & 0.536 & 0.516 & 0.446 & 0.376 & 0.381 \\
     & \checkmark & \checkmark & \checkmark & 0.536 & \textbf{0.517} & 0.447 & 0.375 & \textbf{0.387} \\
    \midrule
     & & & Time point & & \\
    \midrule
    \multirow{6}*{TL} & \checkmark & & Single & 0.504 & 0.450 & 0.478 & 0.217 & 0.320 \\
     & \checkmark & & Multiple & 0.537 & 0.482 & \textbf{0.510} & 0.288 & 0.334 \\
     & & \checkmark & Single & 0.127 & 0.208 & 0.067 & 0.134 & 0.176 \\
     & & \checkmark & Multiple & 0.143 & 0.232 & 0.065 & 0.166 & 0.213 \\
     & \checkmark & \checkmark & Single & 0.504 & 0.471 & 0.475 & 0.245 & 0.336 \\
     & \checkmark & \checkmark & Multiple & 0.535 & 0.501 & 0.502 & 0.303 & 0.365 \\
    \midrule
    \multicolumn{4}{l}{BD} & 0.396 & 0.349 & 0.287 & \textbf{0.468} & -\\
    \bottomrule
  \end{tabular}
\end{table*}

\begin{figure}[t]
\centering
\includegraphics[width=0.95\columnwidth]{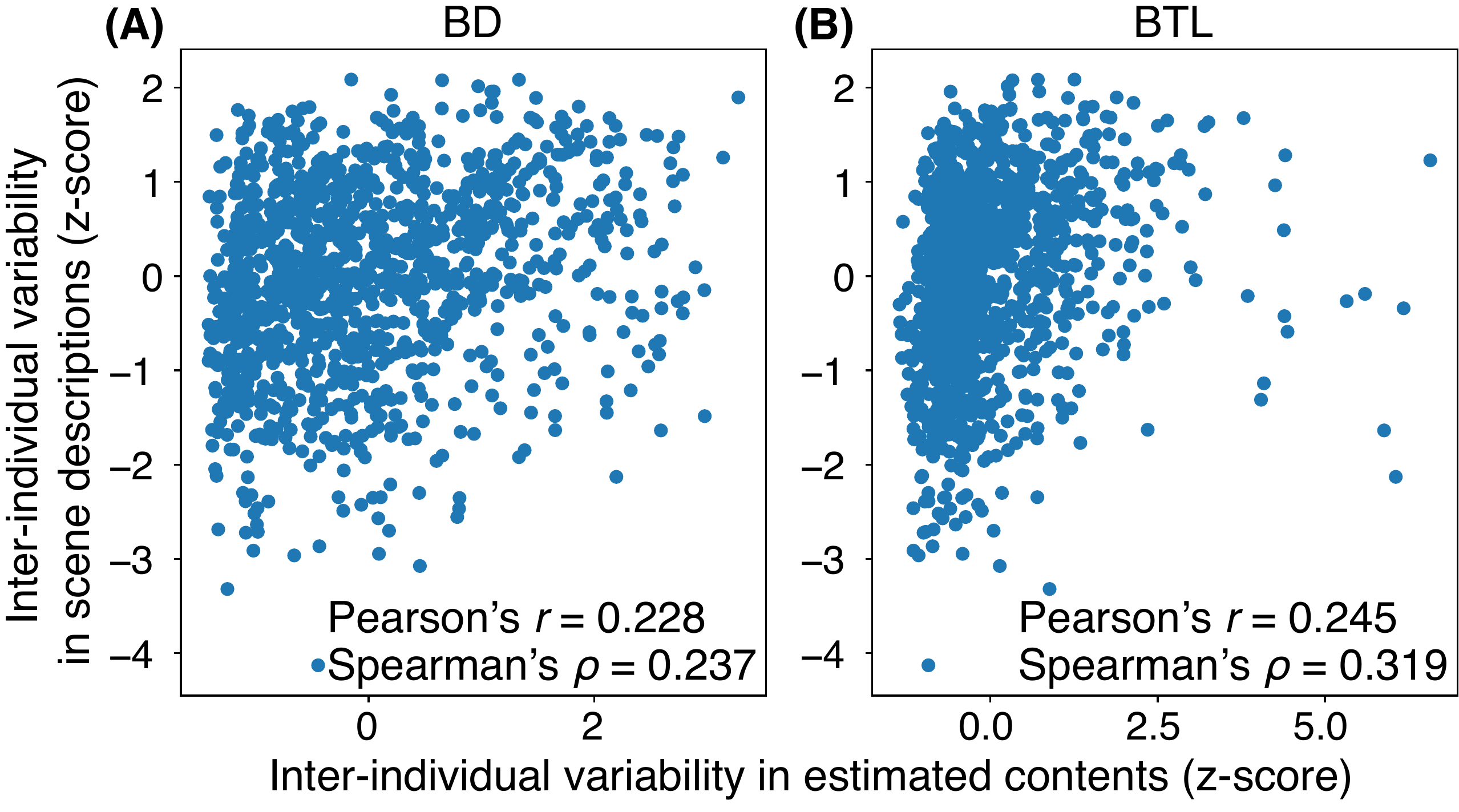} 
\caption{Correlation of individual variability in semantic perception. Each dot depicts the variability in one scene.}
\label{fig:corrindivvar}
\end{figure}

A previous study on brain decoding demonstrated that decoded contents varying brain to brain correlate with individual variability in semantic perception \cite{nishida2018}. To examine whether brain-by-brain variability in contents estimated by BTL has a similar correlation, we evaluated the correlation between individual variability in estimated contents and that in scene descriptions. In this test, BTL with VGG-16 cnn2vox and vox2vox models was used. The individual variability was evaluated by mean pairwise Pearson's correlation distance of estimation- or description-derived word2vec vectors between all possible pairs of brains or annotators. We used a 2-s window slid in 1-s steps to calculate the pairwise distance as is the case with the previous study \cite{nishida2018}. Then, Pearson's and Spearman's correlation coefficients were calculated between the pairwise distances of estimated contents and those of scene descriptions. As a result, there are significant correlations not only for BD, consistent with the previous study \cite{nishida2018}, but also for BTL (Figure \ref{fig:corrindivvar}; t-test, $p < 0.0001$). In addition, the correlation coefficients for BTL (Figure \ref{fig:corrindivvar}B) are rather higher than those for BD (Figure \ref{fig:corrindivvar}A). These results suggest that the models constructed from individual brains through BTL capture the individual variability in semantic perception.

\paragraph{Task2: Estimation of Impression Ratings.}

The second task is the estimation of impression perception induced by each scene of the web ad movies using the impression rating data. The estimation performance is mean values averaged over the 30 impression items. Out of all the methods, the performance of the BTL with the VGG-16 and SoundNet cnn2vox models and the vox2vox model is highest (Table \ref{tab:estpfm}). The best performance in BTL is significantly higher than the best performance in TL and the performance in BD (bootstrapping test, $p < 0.0001$). In this case, in contrast to the estimation of semantic perception, the SoundNet cnn2vox model improves the performance. This indicates that this task requires auditory information and suggests that BTL can extract auditory features from inputs and effectively use them for label estimation.

\paragraph{Task3: Estimation of Ad Effectiveness.}

The third task is the estimation of mass behavior that reflects the effectiveness of each clip of the web ad movies. The effectiveness was evaluated by two behavioral indices collected on the web: click rate (task3-1) and view completion rate (task3-2). For the estimation of click rate, of all the methods, the TL with the VGG-16 layers and the estimation from multiple time points shows the highest performance (Table \ref{tab:estpfm}, task3-1), although its performance is not significantly higher than the best-performing BTL (bootstrapping test, $p = 0.246$). Accordingly, in this case, BTL does not outperform TL. However, this is not surprising because the result that the estimation performance of BD is notably low in this task indicates that brain representations are not efficient for the estimation of click rate. Since BTL guides the estimation through brain representations, the inefficiency of brain representations negatively influences the estimation performance. Indeed, assuming that the vox2vox model makes feature representations more brain-like, the observation that the presence of the vox2vox model reduces the estimation performance may be consistent with this notion.

In contrast to the estimation of click rate, among all the methods, the highest performance in the estimation of view completion is shown by BD (Table \ref{tab:estpfm}, task3-2), although its performance is not significantly higher than the best-performing BTL (bootstrapping test, $p = 0.215$). Following this, BTL shows the second-highest performance despite the fact that TL shows much lower performance. This again indicates that BTL effectively uses brain representations to guide the estimation

\paragraph{Task4: Estimation of Ad Preference.}

The fourth task is the estimation of mass viewers' preference to each clip of the TV ad movies using the ad preference data. In this task, the vox2lab model of BTL was trained using movie datasets completely separated from movie datasets for the training of the cnn2vox and vox2vox models. Hence, the BTL performance in this task reflects how the transformation between CNN and brain features by BTL is effectively generalized to novel datasets. The estimation performance was evaluated using 10-fold cross-validation.

Of all the methods, BTL with the VGG-16 and SoundNet cnn2vox models and the vox2vox model shows the highest performance (Table \ref{tab:estpfm}). The best performance in BTL is significantly higher than the best performance in TL (bootstrapping test, $p < 0.0001$). In this case, the presence of the vox2vox model improves the estimation performance.

\begin{figure}
\centering
\includegraphics[width=0.95\columnwidth]{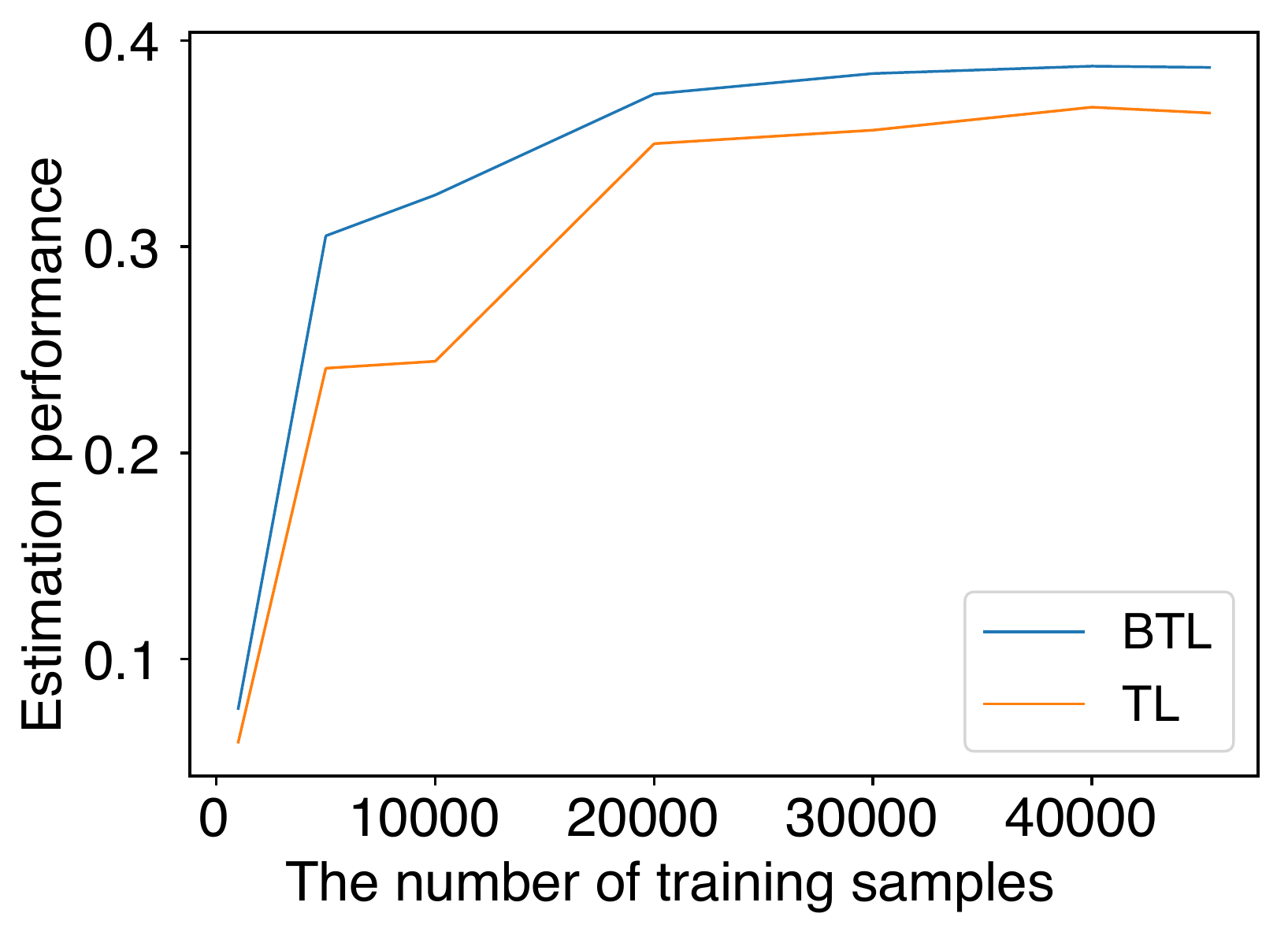} 
\caption{The effect of training sample size on the performance in ad preference estimation. The transition of estimation performance against the training sample size is shown separately for each of BTL (blue) and TL (orange).}
\label{fig:trnsamp}
\end{figure}

In this task, the training of the vox2lab model in BTL and the TL model was conducted on data much larger than the ones used in the other tasks. Hence, to examine the impact of the size of the training data on label estimation, we evaluate the performance of the label estimation by changing the sample size of training data (Figure \ref{fig:trnsamp}). The estimation performance of both the models gradually improves with the increase of the training sample size. Although the performance of each method seems to reach a plateau when all the samples were used, BTL still outperforms TL. This indicates that the superiority of BTL is not due to the difference between BTL and TL in terms of the sample size required for model training.

\section{Conclusion}

We proposed BTL for guiding the estimation of cognitive labels from complex audiovisual inputs through the transformation of CNN features to brain representations. To evaluate the performance of BTL and compare it with the performance of the other methods, we conducted four different types of recognition tasks. Our results in these recognition tasks demonstrated that BTL outperforms regular TL especially in the tasks in which brain representations are effective for the estimation. In addition, our results also suggest that the BTL estimation captures individual variability in perception. Therefore, BTL with CNNs potentially provides a powerful method to improve the generalization ability of CNNs in the estimation of human-like cognition and behavior, including individual differences.

The modeling in BTL is based on building voxelwise encoding models \cite{naselaris2011}. In this line of research, there are many ongoing attempts to enhance modeling performance in terms of brain-response prediction \cite{celik2018,nunez-elizalde2019,wen2016}. Such enhancement potentially enables to utilize brain feature representations more efficiently by BTL, resulting in better performance in the estimation of human-like cognition and behavior, including their individual differences. In addition, the voxelwise modeling basically allows the prediction of brain response from any sort of feature representations. This means that BTL can also be applied to other machine-learning methods different from CNNs. Therefore, BTL has the potential to improve the generalization ability of overall machine-learning methods. We believe that BTL makes a key contribution to further developments in the field of machine learning.

\subsubsection*{Acknowledgments}

The work was supported by grants from the Japan Society for the Promotion of Science (JSPS; KAKENHI 18K18141 and 17H01797) and Tateishi Science and Technology Promotion Foundation (2191025) to S. Nishida, from JSPS (JP15H05311) and ERATO JPMJER1801 to S. Nishimoto, and from NTT DATA Corporation to S. Nishida and S. Nishimoto. We also thank Mr. Koji Takashima, Mr. Takeshi Matsuda, Mr. Susumu Minamiyama, and Ms. Mami Yamashita for their analytical and experimental supports. We also thank Mr. Ryo Yano and Ms. Risa Matsumoto, who are employees of NTT DATA Corp., and Mr. Masato Okino and Mr. Akira Nagaoka, who are employees of NTT DATA Institute of Management Consulting, Inc., for providing movie materials/labels and helping fMRI data collection.

\bibliography{Nishida-BTL-AAAI}
\bibliographystyle{aaai}

\end{document}


\makeatletter
\renewcommand{\figurename}{Supplementary Figure}
\renewcommand{\tablename}{Supplementary Table}
\makeatother

\section*{Supplementary Methods}

\subsection*{fMRI Data Collection and Preprocessing}

fMRI responses to the movies were collected from Japanese participants with normal or corrected-to-normal vision. Forty participants (age 20--61, mean = 26.6; 15 females) were assigned to fMRI experiments with the web ad movies, and 28 participants (age 20--48, mean = 26.4; 12 females) were assigned to fMRI experiments with the TV ad movies. Sixteen participants were overlapped between these two experiments. A 3T Siemens MAGNETOM Prisma scanner (Siemens, Germany) was used with a 32-channel Siemens volume coil and a multiband gradient echo-EPI sequence \cite{moeller2010} (TR = 1000 ms; TE = 30 ms; flip angle = $60^\circ$; voxel size = $2 \times 2 \times 2$ mm; matrix size = 96 $\times$ 96; FOV = $192 \times 192$; the number of axial slices = 72; multiband factor = 6). The field of view covered the entire cortex. Using a T1-weighted MPRAGE sequence on the same 3T scanner, anatomical data were also collected (TR = 2530 ms; TE = 3.26ms; flip angle = $9^\circ$ ; voxel size = $1 \times 1 \times 1$ mm; matrix size = $256 \times 256$; FOV = $256 \times 256$ mm). The experimental protocol is approved by the ethics and safety committees of the National Institute of Information and Communications Technology (NICT).

In the experiments, the participants viewed the movie stimuli on a projector screen inside the scanner ($27.9^\circ \times15.5^\circ$ of visual angle) and listened to the accompanying sounds through MR-compatible headphones. The participants were given no explicit task. The fMRI response data for each of the movies were collected from individual participants in three separate recording sessions over 3 days.

To create the movie stimuli, the original movies in each of the web and TV ad datasets were sequentially concatenated in a pseudo-random order. For each dataset, 14 non-overlapping movie clips of 610 s in length were obtained. The individual movie clips were displayed in separate scans. The initial 10-s part of each clip is a dummy to discard hemodynamic transients caused by clip onset. The fMRI responses collected during the 10-s dummy movie are not used for the modeling. The 12 clips including movies of the training dataset were presented once. The fMRI responses to these clips are used for model training (7200 s in total). The other clips including movies of the test dataset were presented four times each in four separate scans. The fMRI responses to these clips were averaged across four scans and used for model test (1200 s in total). Note that for the TV ad dataset, the 420 movies presented in the fMRI experiments are used for the training of the cnn2vox and vox2vox models, whereas the remaining 2032 movies are used for the training of the vox2lab model and the test of its estimation performance.

For fMRI data preprocessing, motion correction in each functional scan is performed using the statistical parameter mapping toolbox (SPM8, \url{http://www.fil.ion.ucl.ac.uk/spm/software/spm8/}). For each participant, all volumes are aligned to the first image from the first functional run. Using a median filter with a 120-s window and subtracted from the signal, low-frequency fMRI response drift is detected. Then, the response for each voxel is normalized by subtracting the mean response and scaling it to the unit variance. FreeSurfer \cite{dale1999,fischl1999} is used to identify cortical surfaces from anatomical data and register them to the voxels of functional data. All voxels identified within the whole cortex for each participant are used for the modeling.

\subsection*{Details of the Cognitive Labels}

\paragraph{Scene Description Data.}

Manual scene descriptions using natural Japanese language were given for every 1-s scene of the web ad movies. The annotators were native Japanese speakers (2 males and 11 females; age 20--56) who were not the fMRI participants. They were instructed to describe each scene using more than 50 Japanese characters. Five annotators were randomly assigned for each scene to reduce the potential effect of personal bias. The descriptions contain a variety of expressions reflecting not only their objective perceptions but also their subjective perceptions (e.g., impression, feeling, association with ideas). The following are the descriptions of two example scenes:

\vspace{\baselineskip}

\noindent Example scene 1
\begin{itemize}
  \item In the middle of the light blue background, there is a man who has a tennis racket in his hand and glares at the front. The scene tells me he is very serious.
  \item He is the tennis player who everyone really admires. He has a cool face. He is handsome and refreshing. It looks like an advertisement for men.
  \item A short-haired man with a suntanned skin looks at a point with a sharp look. He has a racket in his hand. He is wearing blue clothes.
  \item The power of his eyes is amazing. His sharp eyes give me the impression that he intimidates his opponent. As a sportsman, he seems to be clean even if he is sweating.
  \item A young Asian man with short black hair faces his eyes forward. I can see his racket. There are white letters in the lower part of the screen.
\end{itemize}

\vspace{\baselineskip}

\noindent Example scene 2
\begin{itemize}
  \item It looks like a Christmas scene. It seems like Santa Claus is going back to the sky. It looks good for a Christmas gift.
  \item A book on the green cover is opened. The book is on the brown table. In the book, there is a round green character waving its hands.
  \item A thick open book with a green spine cover is placed on the table of brown trees. It looks like a very funny picture book.
  \item A green book is opened on the desk of the tree. On the opened page, a moon is drawn with a round green stuffed animal waving toward the moon.
  \item It is cute. It seems to be waving bye-bye. Before Christmas? It seems like he is looking down on Mr. Reindeer and Santa.
\end{itemize}

\vspace{\baselineskip}

To evaluate the scene descriptions quantitatively, the descriptions were transformed into vectors of word2vec \cite{mikolov2013} in the same way as in a previous work \cite{nishida2018}. The word2vec (skip-gram) algorithm was originally developed to learn a high-dimensional word vector space based on local (nearby) word co-occurrence statistics in natural language texts. A word2vec vector space was constructed ahead from a text corpus of the Japanese Wikipedia dump on January 11, 2016 (vector dimensionality = 100; window size = 10; the number of negative samples = 5). All Japanese texts in the corpus were segmented into words and lemmatized using MeCab (\url{http://taku910.github.io/mecab}). The only parts of speech used were nouns, verbs, and adjectives, and the others were discarded. To improve the reliability of the word2vec learning and restrict the vocabulary size to around 100000 words, words that appeared less than 178 times in the corpus are excluded.

Each description for a given scene was also segmented, lemmatized, and decomposed into nouns, verbs, and adjectives using MeCab as described above. Individual words were projected into the word2vec vector space. The word vectors were averaged within each description. Then, for each scene, all vectors that were obtained from the different descriptions are averaged. This procedure yielded one 100-dimensional vector (description vector) for each 1-s scene.

\paragraph{Impression Rating Data.}

Subjective ratings of 30 different impressions were manually conducted by human annotators (6 males and 26 females; age 24--61) who were not the fMRI participants. The ratings were given for every 2-s scene of the web ad movies. The impression items were ''beautiful,'' ''ugly,'' ''cute,'' ''disgusting,'' ''urban,'' ''country,'' ''luxury,'' ''cheap,'' ''womanly,'' ''manly,'' jolly,'' ''dismal,'' ''contemporary,'' ''traditional,'' ''simple,'' ''complex,'' ''warm,'' ''cool,'' ''delicate,'' ''bold,'' ''intelligent,'' ''funny,'' ''mild,'' ''dynamic,'' ''humanly,'' ''mechanical,'' ''clean,'' ''filthy,'' ''noisy,'' and ''quiet.'' While the annotators sequentially watched 2-s separate clips of the movies with sounds, they evaluated each item on a scale of 0 to 4. To keep the motivation of the annotators, only 15 labels were assigned to each annotator in single clip evaluation. All the items were evaluated from nine different annotators for each scene. The mean impression ratings in every 2-s scene were obtained by averaging the nine ratings and then oversampled to obtain time series of rating labels in every 1-s scene.

\paragraph{Ad Effectiveness Index Data.}

To evaluate the effectiveness of individual web ad movies, two types of mass behavior indices were collected on the web. One index is click rate, that is the fraction of viewers who clicked the frame of a movie and jumped to a linked web page. The other index is view completion rate, that is, the fraction of viewers who continued to watch an ad movie until the end without choosing a skip option. These indices were computed from 3783 to 12268028 (mean = 632516) unique accesses for each movie. Although a single value of each index was assigned to each movie clip, time series of indices in every 1-s scene were obtained by filling all 1-s scenes in a clip with an identical index value assigned to the clip.

\paragraph{Ad Preference Data.}

Mass preference for each clip in the TV ad movies was collected for commercial investigation using questionnaires conducted on large-scale testers. In the questionnaires, a tester was asked to freely recall a small number of her/his favorite TV ads for him/her from among the ads recently broadcasted on TV. To eliminate the bias of preference due to frequent exposure on TV, the number of recalls of an ad was divided by its gross rating point (GRP). GRP is an index of how many people see an ad over a particular time period, which is calculated by the audience ratio multiplied by the number of times the ad is broadcasted during the period. Although a single preference value was assigned to each movie clip, time series of the value in every 1-s scene were obtained by filling all 1-s scenes in a clip with an identical value assigned to the clip.

\bibliographystyle{aaai}
\bibliography{Nishida-BTL-AAAI-Supplementary}

\onecolumn

\section*{Supplementary Tables and Figures}

\begin{table}[h]
  \caption{The number of movie clips belonging to individual product/service categories}
  \centering
  \begin{tabular}{lll}
    \toprule
    Categories & Web ad movies & TV ad movies \\
    \midrule
    Electronic \& Precision & 4 & 50 \\
    Audiovisual & 5 & 6 \\
    Appliance & 16 & 23 \\
    Car & 31 & 145 \\
    Food \& Confectionery & 7 & 369 \\
    Beverage \& Alcoholic drink & 20 & 236 \\
    Medical \& Health & 35 & 156 \\
    Cosmetics & 49 & 85 \\
    Sundries \& Home equipment & 10 & 254 \\
    Garment/apparel & 9 & 43 \\
    Entertainment & 42 & 237 \\
    Media \& Education & 41 & 82 \\
    Distribution \& Retailer & 12 & 112 \\
    Communication \& Service & 35 & 328 \\
    House \& Construction & 9 & 90 \\
    Finance & 9 & 145 \\
    Enterprise, Public service, \& Others & 34 & 91 \\
    \bottomrule
  \end{tabular}
\end{table}

\vspace{5em}

\begin{figure}[h]
  \centering
  \includegraphics[width=0.95\textwidth]{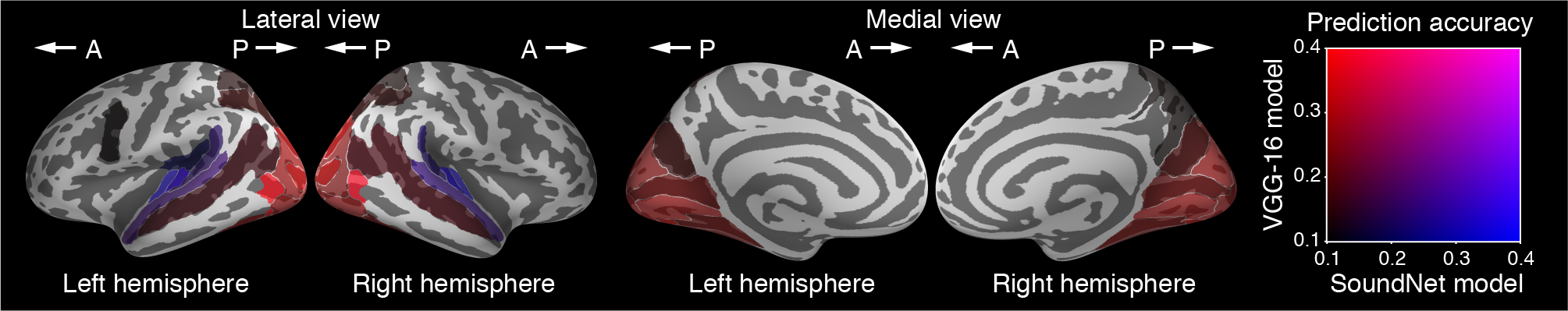}
  \caption{Localization of cortical regions, the response of which is predictable by each cnn2vox model. The cortex of each participant was anatomically segmented into 150 region-of-interests (ROIs) on the basis of the Destrieux atlas \cite{destrieux2010}. Then, population-averaged prediction accuracy within each ROI was mapped onto the cortical surface of a reference brain. Colors indicate the mean prediction accuracy within each of the ROIs denoted by the colormap (right). The accuracy of less than 0.1 is not shown. The ROIs predicted preferentially by the VGG-16 model sets and the ROIs predicted preferentially by the SoundNet model sets are shown in different colors (red and blue, respectively). The VGG-16 models accurately predict voxel responses in occipital and posterior temporal cortical areas, in which visual areas are located. In contrast, the SoundNet models accurately predict voxel responses in anterior temporal cortical areas, in which auditory areas are located; A, anterior; P, posterior.}
  \label{fig:predaccmap}
\end{figure}